\documentclass{article}
\usepackage{ijcai26}

\usepackage{times}
\usepackage{soul}
\usepackage{url}
\usepackage[hidelinks]{hyperref}
\usepackage[utf8]{inputenc}
\usepackage[small]{caption}
\usepackage{graphicx}
\usepackage{amsmath}
\usepackage{amsthm}
\usepackage{amssymb}
\usepackage{booktabs}
\usepackage{algorithm}
\usepackage{algpseudocode} 
\usepackage{multirow}
\usepackage{xcolor}
\usepackage[switch]{lineno}

\title{PRISM: Probability Reallocation with In-Span Masking for \\Knowledge-Sensitive Alignment}

\author{
Chenning Xu
\and
Mao Zheng\and
Mingyang Song\\
\affiliations
Large Language Model Department, Tencent, China\\
\emails
\{chenningxu, moonzheng, nickmysong\}@tencent.com
}

\begin{document}

\maketitle

\begin{abstract}
Supervised fine-tuning (SFT) with token-level hard labels can amplify overconfident imitation of factually unsupported targets, causing hallucinations that propagate in multi-sentence generation. We study an augmented SFT setting in which training instances include coarse sentence-level factuality risk labels and inter-sentence dependency annotations, providing structured signals about where factual commitments are weakly supported. We propose \textbf{PRISM}, a differentiable risk-gated framework that modifies learning only at fact-critical positions. PRISM augments standard SFT with a lightweight, model-aware probability reallocation objective that penalizes high-confidence predictions on risky target tokens, with its scope controlled by span-level risk weights and model-aware gating. Experiments on hallucination-sensitive factual benchmarks and general evaluations show that PRISM improves factual aggregates across backbones while maintaining a competitive overall capability profile. Ablations further show that the auxiliary signal is most effective when used conservatively, and that knowledge masking and model-aware reallocation play complementary roles in balancing factual correction and capability preservation.
\end{abstract}

\begin{figure}[t]
    \centering
    \includegraphics[width=0.85\linewidth]{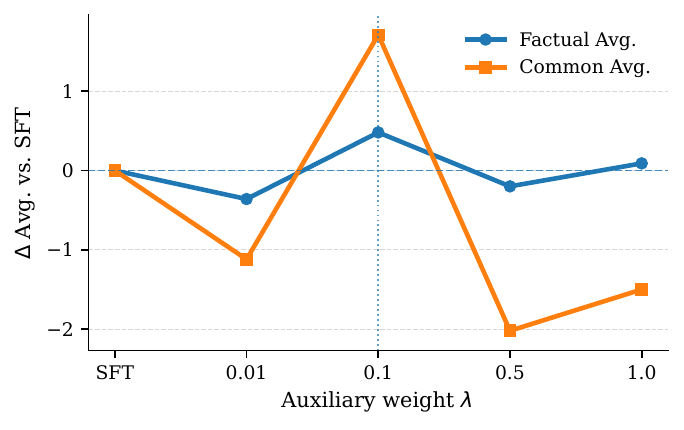}
    \caption{Averaged factual and common performance under different auxiliary weights $\lambda$, shown as $\Delta$ relative to the SFT baseline. The figure illustrates a trade-off between factual improvement and preservation of general capability.}
    \label{fig:intro_delta_avg}
\end{figure}

\section{Introduction}

Large language models (LLMs) trained with next-token prediction can produce fluent text and follow instructions after supervised fine-tuning (SFT). In practice, their usability on downstream applications is often driven by SFT on instruction data and, in some settings, further preference-based alignment. Instruction tuning and related pipelines substantially improve adherence to user intent and overall helpfulness, making SFT a standard component of modern LLM training recipes \cite{wei2022finetuned,ouyang2022training}.

Despite these gains, factual hallucination remains a persistent failure mode: models may produce fluent statements that are unsupported by the given context, internally inconsistent with earlier claims, or simply false. This has been documented across summarization \cite{maynez-etal-2020-faithfulness}, open-domain question answering and truthfulness benchmarks \cite{lin-etal-2022-truthfulqa}, and long-form generation settings where errors can compound over multiple sentences \cite{min-etal-2023-factscore,wei2024longform}. Surveys further highlight that hallucination is shaped by both modeling choices and data/optimization pipelines, and that mitigation strategies vary widely across tasks and assumptions \cite{ji2023survey,huang2025survey}.

A broad set of approaches has been explored to reduce hallucinations. Retrieval-augmented generation (RAG) grounds outputs in external corpora and can improve factuality by conditioning on retrieved evidence \cite{lewis2020rag}. Post-hoc or self-consistency style verification aims to detect unsupported claims after generation, e.g., by sampling and checking consistency \cite{manakul-etal-2023-selfcheckgpt}. Preference-based optimization methods provide another training signal to steer model behavior (often without explicit fact annotations), with objectives such as DPO offering a simpler alternative to RLHF-style pipelines \cite{rafailov2023dpo}. While effective in many settings, these solutions typically introduce additional components (retrieval, verifiers, or multi-stage pipelines) and they do not directly address a basic tension inside SFT: maximizing likelihood on imperfect training targets can also reinforce unsupported or erroneous factual statements.

This paper studies a training-time alternative that keeps the simplicity of SFT while injecting structured factual signals into the optimization process. We consider instruction datasets augmented with (a) extracted atomic facts (minimal factual propositions) and (b) a relational structure capturing cross-sentence factual support and dependencies. Atomic-fact decompositions are increasingly used to evaluate long-form factuality and to localize errors at a finer granularity \cite{min-etal-2023-factscore,wei2024longform}. Our goal is not to convert SFT into full verification, but to use these annotations to identify \emph{fact-critical} positions where the model introduces new commitments or relies on unsupported links across sentences.

Concretely, we propose PRISM, a fact-aware SFT framework that aims to balance two desiderata: (1) learning general instruction-following behavior from diverse SFT data and (2) reducing unsupported factual claims in hallucination-sensitive settings such as fact-seeking QA. PRISM couples standard cross-entropy with additional, selectively applied objectives that reallocate token-level probability mass on fact-critical spans, discouraging high-confidence realizations of unsupported atomic facts while avoiding uniform suppression on unrelated tokens. The update is conditioned on the model's current preference, which targets compensation to tokens the model is most likely to over-commit to, and a single coefficient controls the strength of this compensation. This design is aligned with recent evidence that SFT-time interventions---including data filtering, abstention-style learning, and confidence-aware objectives---can materially affect hallucination behavior \cite{si-etal-2025-aligning,huang-etal-2025-alleviating,yuan-etal-2025-beyond,huang-chen-2024-factalign,nguyen2025smoothing}.

We evaluate PRISM on both hallucination-sensitive factual benchmarks and general capability benchmarks to quantify whether factual improvements come at the cost of broader instruction-following and reasoning performance. Across backbones, PRISM yields consistent gains in factual aggregates relative to vanilla SFT while maintaining a competitive general profile, and ablations characterize how the compensation weight and model-aware update rule influence this balance.

Our contributions are threefold:
\begin{itemize}
    \item We formulate knowledge-aware SFT using atomic facts and fact-relation structure extracted from instruction data, providing factual signals without requiring external retrieval at training time.
    \item We propose PRISM, a selective probability-reallocation framework that conditions updates on fact-critical spans and the model's current preference to control overconfident factual commitments during SFT.
    \item We provide empirical analysis on both general and hallucination-focused evaluations, including ablations that isolate the effects of compensation strength and model-aware updates on factuality--capability interactions.
\end{itemize}

\section{Related Work}

\subsection{Hallucination Mitigation}

Hallucinations in large language models (LLMs) are commonly addressed via system- or inference-level mechanisms that improve grounding or add external feedback. Retrieval-augmented generation injects evidence at generation time to reduce unsupported statements \cite{lewis2020rag}, while post-hoc verification and self-checking aim to detect, critique, or filter hallucinated content without modifying the base model \cite{manakul2023selfcheckgpt}. Preference-based alignment provides another axis of control by optimizing models toward human-preferred behavior; this can indirectly reduce hallucinations, but typically introduces additional preference data and training stages \cite{rafailov2023dpo}. Broad surveys organize these directions and emphasize recurring tensions among factuality, helpfulness, and efficiency \cite{huang2025survey}.

Recent work further targets factuality with fine-grained supervision signals and knowledge-boundary modeling. FactAlign improves long-form factual consistency using sentence-level alignment signals derived from automatic factuality assessment \cite{huang-chen-2024-factalign}. APEFT constructs atomic-statement preferences to address under-alignment and improve factual robustness under distribution shift \cite{yuan-etal-2025-beyond}. UAlign leverages uncertainty estimates (e.g., confidence and entropy) to represent knowledge boundaries and uses them within alignment objectives \cite{xue-etal-2025-ualign}. While effective, these pipelines often depend on evaluators, preference construction, uncertainty estimation, or reward modeling components, motivating simpler training-time interventions that integrate into standard supervised fine-tuning.

\subsection{Hallucination Mitigation during SFT}

A growing line of work argues that hallucinations can be introduced or amplified during supervised fine-tuning (SFT) due to \emph{knowledge misalignment}: hard-label imitation can push models toward confident reproduction of targets that exceed their reliable knowledge. Controlled analyses show that introducing unfamiliar or newly added knowledge during fine-tuning increases hallucination tendencies as such examples are eventually learned \cite{gekhman-etal-2024-fine}. Complementarily, studies on factual knowledge extraction highlight failure modes where fine-tuning on poorly stored (less salient) facts can suppress subject-specific cues, leading to generic yet plausible answers even when relevant knowledge is encoded \cite{ghosal2024understanding}. Together, these findings suggest that SFT-time interventions should account for both (i) whether supervision is aligned with the model's pretraining knowledge and (ii) how training signals shape confidence on fact-critical spans.

To mitigate hallucinations during SFT, recent methods intervene through objective design and data selection. SEAL introduces selective abstention learning, enabling the model to reject misaligned supervision rather than over-commit to incorrect tokens \cite{huang-etal-2025-alleviating}. Data-centric approaches such as NOVA reduce hallucinations by filtering instruction-tuning data based on the model's familiarity with candidate samples, aiming to preserve instruction-following while avoiding unfamiliar targets \cite{si-etal-2025-aligning}. Other objective-shaping techniques reduce overconfidence by replacing one-hot supervision with softer targets, including smoothed knowledge distillation \cite{nguyen2025smoothing}. FLAME further studies factuality-aware alignment across both SFT and post-training, illustrating how standard alignment objectives can inadvertently encourage hallucinations and proposing factuality-aware alternatives \cite{lin2024flame}. Our work follows this training-time direction: rather than relying on retrieval or decoding-time constraints, it modifies SFT objectives to control confidence updates on knowledge-sensitive tokens, and our ablations examine how auxiliary strength and update rules affect the balance between factual benchmarks and general instruction-following.

\section{Methodology}
\label{sec:method}

\begin{figure*}[t]
  \centering
  \includegraphics[width=0.62\textwidth]{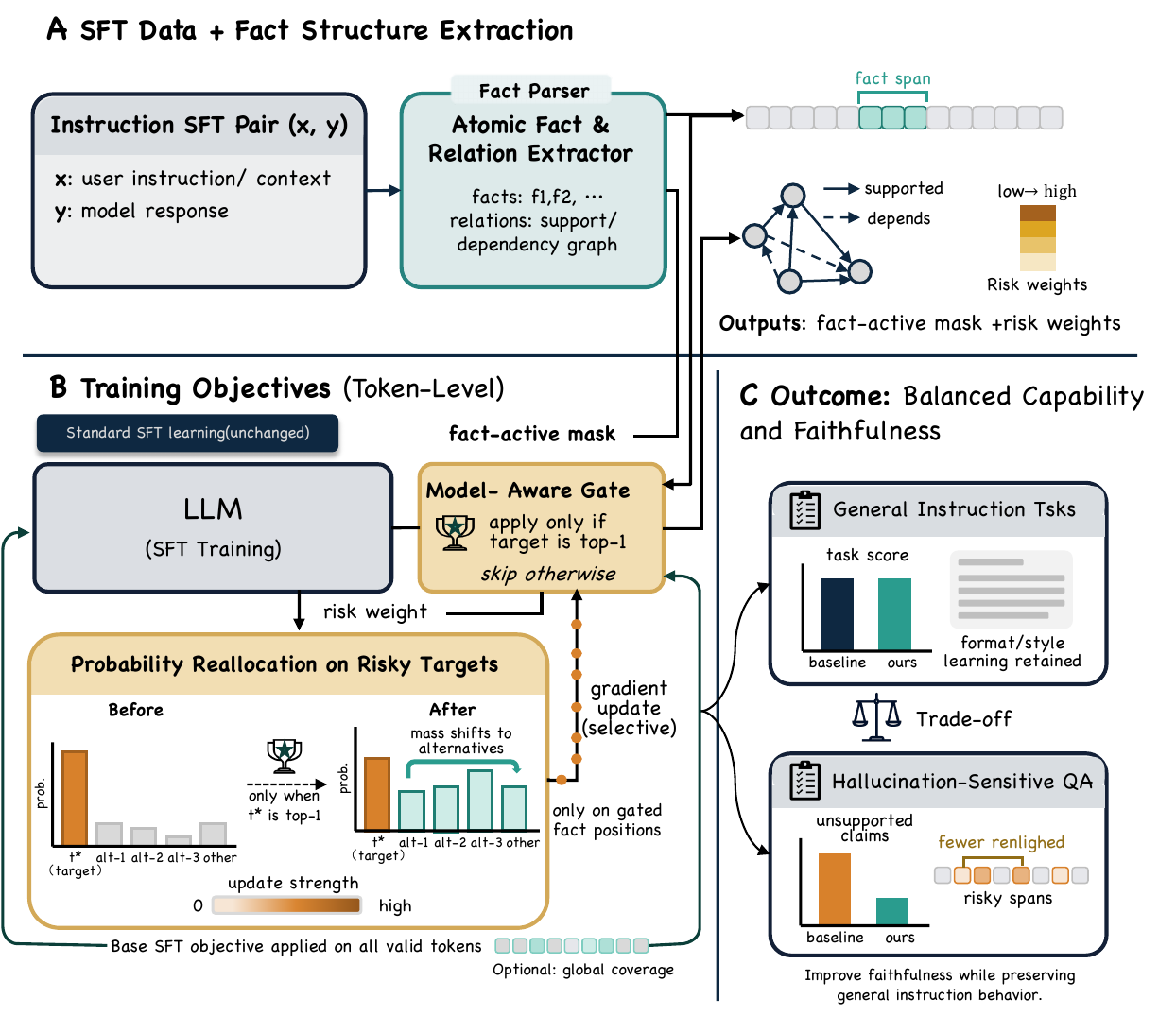}
  \caption{Overall pipeline of our risk-gated knowledge-aware SFT (PRISM). Starting from instruction--response pairs, we extract atomic facts and fact relations, derive token-level fact spans and risk weights, and inject these signals into SFT via a model-aware, risk-gated probability reallocation objective.}
  \label{fig:overview}
\end{figure*}

\subsection{Setup and Notation}
We consider single-turn SFT examples $(x, y)$, where $x$ is the user input (instruction plus optional context) and $y=(y_1,\ldots,y_T)$ is the target response.
An autoregressive model parameterized by $\theta$ produces logits $\mathbf{z}_t \in \mathbb{R}^{|\mathcal{V}|}$ and token probabilities
$p_\theta(v \mid x, y_{<t}) = \mathrm{softmax}(\mathbf{z}_t)_v$ over vocabulary $\mathcal{V}$.
We train on valid positions indicated by a mask $\mathbf{m}^{\text{valid}}_t \in \{0,1\}$ (e.g., excluding padding and non-target regions).
For a response $y$, we further segment it into sentences $\{s_j\}_{j=1}^{J}$ (by standard punctuation rules), and denote by $\mathrm{sid}(t)\in\{1,\dots,J\}$ the sentence index containing token $y_t$.

\subsection{Fact-Aware Signals from SFT Data}
Given each target response $y$, we derive a coarse factual structure consisting of (i) a set of \emph{atomic facts} $\mathcal{F}=\{f_i\}_{i=1}^{M}$ (minimal factual propositions) and (ii) a set of directed \emph{fact relations} $\mathcal{E}$ capturing cross-sentence support/dependency.
Each atomic fact $f_i$ is aligned to one or more token spans in $y$; these spans mark where the response makes a concrete factual commitment (e.g., entities, numbers, and their predicates).
We map this structure to token-level training signals used by PRISM.

\paragraph{Fact-active mask.}
We define a \emph{fact-active mask} $\mathbf{m}^{\text{fact}}_t \in \{0,1\}$ that indicates whether position $t$ belongs to any fact-aligned span.
The mask is intentionally coarse: it does not require token-level corrections, only the identification of fact-critical spans where factual commitments are introduced.

\paragraph{Risk weights with dependency propagation.}
In addition, each sentence $s_j$ is assigned a coarse factuality-risk score $r_j\in[0,1]$ (higher means more risky / less supported) and sentence-level dependency edges in $\mathcal{E}$ indicate that $s_j$ relies on factual content introduced in some prior sentence(s).
To reflect that unsupported claims can propagate through dependencies, we construct an \emph{effective} sentence risk
\begin{equation}
\tilde r_j \;=\; \max\Big(r_j,\; \max_{(j'\rightarrow j)\in \mathcal{E}} r_{j'}\Big),
\label{eq:risk_prop}
\end{equation}
i.e., a sentence inherits risk from its immediate supporting sentences.
We then map sentence risk to a per-token \emph{support weight} $w_t\in[0,1]$ by
\begin{equation}
w_t \;=\; 1-\tilde r_{\mathrm{sid}(t)},
\label{eq:wt}
\end{equation}
so that larger $w_t$ indicates stronger support (lower risk) and smaller $w_t$ indicates weaker support (higher risk).
This design matches our supervision setting: the annotations provide structured but coarse guidance about which spans/sentences are fact-sensitive and how risk relates across sentences, without requiring explicit alternative targets.

\subsection{Training Objective}
\label{sec:training_objective}

\paragraph{Standard SFT loss.}
We optimize the usual token-level negative log-likelihood over valid positions, where $N_{\text{sft}}=\sum_t \mathbf{m}^{\text{valid}}_t$:
\begin{equation}
\mathcal{L}_{\text{SFT}}
=
-\frac{1}{N_{\text{sft}}}
\sum_{t=1}^{T}
\mathbf{m}^{\text{valid}}_t
\log p_\theta(y_t \mid x, y_{<t})
\label{eq:sft}
\end{equation}

\paragraph{Risk-gated probability mass reallocation.}
On fact-critical spans, we add an auxiliary objective that discourages overly confident probability mass on the ground-truth token when it is deemed risky.
Let $p_t^{\text{label}} = p_\theta(y_t \mid x, y_{<t})$ denote the model probability assigned to the label token at step $t$, and let
\begin{equation}
q_t = \max_{v \neq y_t} p_\theta(v \mid x, y_{<t})
\end{equation}
be the largest competing probability.

A naive top-1 gate is not sufficient here: if probability mass is reallocated away from the label token so aggressively that it is no longer the preferred token, then the auxiliary update no longer acts on the current factual target, but effectively shifts optimization toward an alternative continuation. This creates a mismatch between the intended factual correction and the sequence actually favored after redistribution.

To avoid this, we use a \emph{two-stage model-aware gate}. First, we require that the label token is currently preferred:
\begin{equation}
\mathbf{g}^{\text{pref}}_t
=
\mathbf{I}\Big[
y_t = \arg\max_{v \in \mathcal{V}} p_\theta(v \mid x, y_{<t})
\Big].
\label{eq:pref_gate}
\end{equation}

Second, we require that after applying the intended probability reallocation, the label token would still remain the top prediction.
Given reallocation weight $w_t \in [0,1]$, we consider the induced redistribution
\begin{equation}
p_t^{\text{label}} \rightarrow p_t^{\text{label}} w_t,
\qquad
p_t(v) \rightarrow p_t(v)\cdot \frac{1 - p_t^{\text{label}} w_t}{1 - p_t^{\text{label}}}
\quad (v \neq y_t),
\label{eq:redistribution}
\end{equation}
which preserves normalization.
The label remains preferred after redistribution iff
\begin{equation}
p_t^{\text{label}} w_t
\;\ge\;
q_t \cdot \frac{1 - p_t^{\text{label}} w_t}{1 - p_t^{\text{label}}},
\label{eq:still_top_raw}
\end{equation}
or equivalently,
\begin{equation}
p_t^{\text{label}} w_t (1 - p_t^{\text{label}})
\;\ge\;
q_t (1 - p_t^{\text{label}} w_t).
\label{eq:still_top}
\end{equation}
We define the corresponding consistency gate:
\begin{equation}
\mathbf{g}^{\text{keep}}_t
=
\mathbf{I}\Big[
p_t^{\text{label}} w_t (1 - p_t^{\text{label}})
\ge
q_t (1 - p_t^{\text{label}} w_t)
\Big].
\label{eq:keep_gate}
\end{equation}

The final auxiliary weight is
\begin{equation}
\alpha_t
=
\mathbf{m}^{\text{fact}}_t
\cdot
\mathbf{g}^{\text{pref}}_t
\cdot
\mathbf{g}^{\text{keep}}_t
\cdot
(1 - w_t).
\label{eq:alpha_new}
\end{equation}
Thus, high-risk positions (small $w_t$) receive stronger penalties, but only when the update remains aligned with the current factual target before and after redistribution.

We use a \emph{complement} loss on the label probability:
\begin{equation}
\mathcal{L}_{\text{comp}}
=
\frac{1}{N_{\text{fact}}}
\sum_{t=1}^{T}
\alpha_t
\left(
-\log\left(1 - \tilde p_t^{\text{label}}\right)
\right),
\quad
N_{\text{fact}}=\sum_t \mathbf{m}^{\text{fact}}_t,
\label{eq:comp}
\end{equation}
where $\tilde p_t^{\text{label}}=\min(p_t^{\text{label}},\,1-\epsilon)$ clamps the probability for numerical stability (equivalently, using $\textit{log1p}$).
Intuitively, minimizing $\mathcal{L}_{\text{comp}}$ decreases $p_t^{\text{label}}$ at selected positions; since token probabilities sum to one, this implicitly reallocates probability mass to alternative tokens according to the model's current distribution.
Crucially, we do not apply this correction whenever it would change the identity of the preferred token after redistribution. The consistency gate in Eq.~\eqref{eq:keep_gate} suppresses such cases, preventing the auxiliary loss from optimizing toward a different continuation than the current factual target.
This reduces mismatch between the intended correction target and the sequence preference induced by the reallocation step.

\paragraph{Final objective.}
Our overall training loss is a weighted sum:
\begin{equation}
\mathcal{L}
=
\mathcal{L}_{\text{SFT}}
+
\lambda \, \mathcal{L}_{\text{comp}},
\label{eq:final_obj}
\end{equation}
where $\lambda$ controls the strength of the auxiliary complement objective.

\subsection{Gradient Analysis}
\label{sec:grad_analysis}

We analyze the gradients induced by the complement objective and their interaction with standard SFT.
At step $t$, let $y=y_t$ and $p_k=p_\theta(k\mid x,y_{<t})$ denote the softmax probabilities from logits $\mathbf{z}_t$.
The per-token complement loss and its normalized objective are
\begin{equation}
\ell_{\mathrm{comp},t} = -\log(1-\tilde p_y),
\qquad 
\mathcal{L}_{\mathrm{comp}}
=
\frac{1}{N_{\mathrm{fact}}}
\sum_t \alpha_t \,\ell_{\mathrm{comp},t},
\label{eq:comp_grad_setup}
\end{equation}
with normalization factor $N_{\mathrm{fact}}$.

Here $\alpha_t$ is treated as constant in backpropagation, since both gates are non-differentiable indicator functions.
Using the derivative of $\ell_{\mathrm{comp},t}$ with respect to $p_y$ together with the softmax Jacobian, we obtain
\begin{equation}
\begin{split}
\frac{\partial \mathcal{L}_{\text{comp}}}{\partial z_y}
&=
\frac{\alpha_t}{N_{\text{fact}}}\,p_y, \\
\frac{\partial \mathcal{L}_{\text{comp}}}{\partial z_k}
&=
-\frac{\alpha_t}{N_{\text{fact}}}\,\frac{p_y}{1-p_y}\,p_k
\quad (k\neq y),
\end{split}
\label{eq:grad_comp}
\end{equation}
and so gradient descent decreases $z_y$ at active positions (reducing $p_y$) and increases other logits in proportion to their current probabilities $p_k$.
This yields a \emph{model-distribution-weighted} probability reallocation: the mass removed from the risky label token is redistributed across alternatives according to the model's present beliefs, without specifying a replacement distribution; the factor $(1-p_y)^{-1}$ makes the correction stronger when the model is highly confident.

However, such redistribution is only useful when it remains aligned with the current factual target. If the reallocation were strong enough to make a competing token overtake the label token, the auxiliary term would effectively optimize toward a different continuation, introducing a target mismatch. Our consistency gate $\mathbf{g}^{\text{keep}}_t$ excludes exactly these cases by requiring that the label token would remain top-1 after the intended redistribution. In practice, we clamp $p_y$ (equivalently, $\tilde p_y$ in $\ell_{\text{comp},t}$) to avoid numerical issues when $p_y$ approaches $1$.

For standard SFT, the gradient encourages the target logit and suppresses alternatives in the usual way.
In contrast, the complement term decreases $z_y$ when activated (Eq.~\eqref{eq:grad_comp}).
Combining this with Eq.~\eqref{eq:final_obj} and the implementation-level masking/normalization, the overall gradient at step $t$ is: 
\begin{equation}
\frac{\partial \mathcal{L}}{\partial z_k}
=
\underbrace{\frac{\mathbf{m}^{\text{valid}}_t}{N_{\text{sft}}}\frac{\partial \ell_{\text{SFT},t}}{\partial z_k}}_{\text{applies on all valid tokens}}
+
\underbrace{\lambda\frac{\partial \mathcal{L}_{\text{comp}}}{\partial z_k}}_{\text{applies only on gated fact positions}},
\label{eq:grad_total_underbrace}
\end{equation}
where $N_{\text{sft}}=\sum_t \mathbf{m}^{\text{valid}}_t$ and the second term is non-zero only when the position is factual, the label token is currently preferred, and it would remain preferred after the intended redistribution (through $\alpha_t$).

\subsection{Training Procedure}
We train with standard teacher-forced next-token prediction.
At each step, $\mathcal{L}_{\mathrm{SFT}}$ is applied to all valid tokens, while $\mathcal{L}_{\mathrm{comp}}$ is applied only to active factual positions through $\alpha_t$.
The two terms are normalized separately by $N_{\mathrm{sft}}$ and $N_{\mathrm{fact}}$, respectively.
Implementation follows the formulation in Section~\ref{sec:training_objective}, with standard numerical safeguards.
All other aspects of SFT remain unchanged, including the optimizer, batching, and decoding-time behavior.
Figure~\ref{fig:overview} summarizes the overall workflow from fact extraction to consistency-gated training.

\section{Experiments}

\subsection{Datasets}

\paragraph{Training Data Source.}
We utilize \textit{lmsys\_chat\_1m\_clean} \cite{openleecher2024lmsyschat1mclean} as our base corpus for supervised fine-tuning (SFT). This dataset is a curated derivative of LMSYS-Chat-1M \cite{zheng2023lmsyschat1m}, comprising one million real-world conversations collected from the Vicuna demo and Chatbot Arena (April--August 2023). The cleaned version filters out non-English items, redacted content, and repetitive spam. Crucially, it replaces the original model outputs with high-quality responses regenerated by \textit{DeepSeek-V3}. We construct single-turn training pairs using the human instruction as the input and the generated assistant response as the target.

\paragraph{Fact-aware Preprocessing.}
To enable fine-grained fact supervision, we process the SFT targets through a three-stage pipeline. First, we segment responses into chunks based on sentence boundaries, imposing a maximum length of 200 tokens per chunk. Second, we extract atomic facts from each chunk and induce directed dependency relations. A relation is established if a fact relies on evidence located in preceding chunks or the user instruction. Finally, we apply a verification filter to discard instances with parsing failures. This results in a set of \textit{Verified Instances} containing span-level annotations used to gate model updates. Table~\ref{tab:data_stats} summarizes the statistics of the processed data.

\subsection{Evaluation}

\paragraph{Factual evaluation.}
We report hallucination detection on \textsc{HHEM} using \textsc{HHEM-2.1-Open}, an open hallucination detection model that scores whether a generation is supported by the provided reference context \cite{hhem21open}.
For knowledge-intensive short-form QA, we report answer accuracy on PopQA \cite{mallen-etal-2023-trust} and TriviaQA \cite{joshi-etal-2017-triviaqa} under a grounded evaluation protocol following Self-RAG \cite{asai2024selfrag}.
Concretely, we adopt the single-retrieval setting for short-form generation and use pre-retrieved documents obtained offline with Contriever \cite{izacard2021contriever}.
For each question, we take the top-5 passages ranked by Contriever as grounding evidence and evaluate answer correctness conditioned on these documents.

\paragraph{Common evaluation.}
To track broad multi-domain knowledge and reduce sensitivity to any single domain, we report accuracy on MMLU (LMSYS split) \cite{hendrycks2021mmlu}, which covers diverse academic and professional subjects and is widely used to monitor overall knowledge retention under post-training. To complement factual QA metrics with capability-oriented evaluations, we additionally include GSM8K \cite{cobbe2021training} for multi-step mathematical reasoning, HumanEval \cite{chen2021evaluating} for functional code generation (pass@1), and IFEval \cite{zhou2023instructionfollowingevaluationlargelanguage} for instruction-following compliance. Together, these benchmarks provide a compact coverage of knowledge, reasoning, coding, and instruction adherence that serves as a general-purpose reference point alongside hallucination-sensitive evaluations.

\subsection{Baselines}
\label{sec:baselines}

We compare PRISM with the following baselines. To assess robustness across model families and scales, we conduct all comparisons on three representative \emph{base} LLM backbones: Qwen3-4B-Base and Qwen3-8B-Base \cite{qwen3technicalreport}, and Llama-3.1-8B \cite{llama3-herd}.

\paragraph{Supervised Fine-Tuning (SFT).}
SFT directly fine-tunes the pre-trained model on the constructed instruction--response pairs, optimizing the standard token-level cross-entropy objective to maximize the likelihood of the ground-truth response conditioned on the input (Eq.~\ref{eq:sft}). This baseline uses the same training data and training recipe as our method, but without any fact-structure extraction or fact-aware auxiliary objectives.

\paragraph{Selective Abstention Learning (SEAL).}
SEAL \cite{huang-etal-2025-alleviating} mitigates hallucinations during SFT by introducing an abstention mechanism for knowledge-misaligned targets. It augments the vocabulary with a special $\textit{[REJ]}$ token and optimizes a training objective that combines standard negative log-likelihood with an additional term that discourages unnecessary abstention when a correct prediction is feasible \cite{huang-etal-2025-alleviating}. SEAL also proposes a decoding-time regularization that uses the model’s step-wise $\textit{[REJ]}$ probability as an uncertainty signal to penalize risky continuations \cite{huang-etal-2025-alleviating}. In our experiments, we adopt SEAL’s training objective and apply the same $\textit{[REJ]}$-based step-wise penalty at inference, while keeping the rest of decoding consistent with our shared evaluation protocol (i.e., without switching to beam search) to avoid search-strength differences as a confounder.

\paragraph{NOVA.}
NOVA \cite{si-etal-2025-aligning} reduces hallucinations during instruction tuning via data filtering based on sample \emph{familiarity} to the backbone model. It scores instruction--response pairs using (i) \emph{Internal Consistency Probing} (agreement among multiple self-generated responses) and (ii) \emph{Semantic Equivalence Identification} (whether the target response is semantically supported by the dominant cluster among model generations, using an NLI-based clustering/voting procedure) \cite{si-etal-2025-aligning}. Following the paper, we generate $K{=}10$ samples per prompt with temperature $T{=}0.7$ for knowledge estimation (Appendix~B in \cite{si-etal-2025-aligning}) and fine-tune on the retained subset selected by NOVA scores. For the optional quality-scoring component, instead of training a task-specific reward model, we use the off-the-shelf reward model Skywork-Reward-Gemma-2-27B-v0.2 \cite{liu-etal-2024-skyworkreward} to score candidate samples, reducing additional reward-model training variance while keeping the filtering pipeline aligned with NOVA’s design.

\paragraph{Knowledge Mask.}
We include a \emph{knowledge mask} baseline to isolate the effect of suppressing updates on potentially unreliable factual spans. Using the same fact-span annotations and per-span risk scores as PRISM, this baseline masks out all fact spans whose factuality score is $<1$ when computing the SFT cross-entropy loss, and optimizes $\mathcal{L}_{\text{SFT}}$ only on the remaining tokens. Unlike PRISM, it does not perform probability reallocation or model-aware update gating, and serves as a controlled ablation-style comparator.

\begin{table*}[t]
\centering
\renewcommand{\arraystretch}{1.12}
\scalebox{0.82}{
\begin{tabular}{lcl|ccccccccc}
\toprule
\multirow{2}{*}{\textbf{Model}} & \multirow{2}{*}{\textbf{Size}} & \multirow{2}{*}{\textbf{Method}} &
\multicolumn{4}{c}{\textbf{Factual}} &
\multicolumn{5}{c}{\textbf{Common}} \\
\cmidrule(lr){4-7}\cmidrule(lr){8-12}
& & &
\textbf{HHEM} $\uparrow$ &
\textbf{PopQA} $\uparrow$ &
\textbf{TriviaQA} $\uparrow$ &
\textbf{Avg} $\uparrow$ &
\textbf{MMLU} $\uparrow$ &
\textbf{GSM8K} $\uparrow$ &
\textbf{HumanEval} $\uparrow$ &
\textbf{IFEval} $\uparrow$ &
\textbf{Avg} $\uparrow$ \\
\midrule

\multirow{4}{*}{Qwen3} & \multirow{4}{*}{4B}
& SFT   & 94.53 & 58.54 & 68.95 & 74.01 & 76.19 & 85.14 & 79.27 & 60.81 & 75.35 \\
& & SEAL  & 92.25 & 57.76 & 68.87 & 72.96 & 76.01 & 87.64 & 79.88 & 60.81 & \underline{76.09} \\
& & NOVA  & 94.83 & 58.89 & 69.34 & \underline{74.35} & 75.28 & 81.58 & 81.71 & 57.86 & 74.11 \\
& & \textbf{PRISM} & 95.03 & 59.54 & 68.90 & \textbf{74.49} & 76.93 & 86.20 & 81.71 & 63.40 & \textbf{77.06} \\
\midrule

\multirow{4}{*}{Qwen3} & \multirow{4}{*}{8B}
& SFT   & 94.83 & 54.90 & 69.22 & 72.98 & 78.21 & 88.02 & 83.54 & 45.66 & 73.86 \\
& & SEAL  & 96.22 & 58.54 & 70.35 & \underline{75.04} & 77.79 & 88.86 & 81.10 & 61.00 & 77.19 \\
& & NOVA  & 94.83 & 59.11 & 70.72 & 74.89 & 76.43 & 90.75 & 82.93 & 61.00 & \textbf{77.78} \\
& & \textbf{PRISM} & 95.13 & 60.83 & 71.22 & \textbf{75.73} & 77.54 & 85.67 & 85.98 & 61.92 & \textbf{77.78} \\
\midrule

\multirow{4}{*}{Llama 3.1} & \multirow{4}{*}{8B}
& SFT   & 88.12 & 58.18 & 68.10 & \underline{71.47} & 50.70 & 51.86 & 40.85 & 45.84 & 47.31 \\
& & SEAL  & 88.57 & 57.76 & 67.94 & 71.42 & 48.62 & 49.73 & 44.51 & 45.10 & 46.99 \\
& & NOVA  & 84.10 & 50.89 & 66.47 & 67.15 & 49.96 & 58.76 & 51.22 & 36.60 & \textbf{49.14} \\
& & \textbf{PRISM} & 89.76 & 59.75 & 69.23 & \textbf{72.91} & 47.72 & 53.30 & 47.80 & 43.07 & \underline{47.97} \\
\bottomrule
\end{tabular}
}
\caption{Main results. \textbf{Factual Avg} is the mean of (HHEM, PopQA, TriviaQA); \textbf{Common Avg} is the mean of (MMLU, GSM8K, HumanEval, IFEval). Best and second-best averages within each backbone are \textbf{bold} and \underline{underlined}, respectively.}
\label{tab:main_results}
\end{table*}

\section{Results}
\subsection{Main Results}

Table~\ref{tab:main_results} indicates that hallucination-aware objectives can change factual reliability (HHEM/PopQA/TriviaQA) and general capability (MMLU/GSM8K/HumanEval/IFEval) in ways that are only partially aligned: improvements on the factual aggregate do not always coincide with gains on the common aggregate, and this coupling differs by backbone. This supports reporting both groups of benchmarks when comparing hallucination-mitigation training, rather than relying on a single summary score.

A consistent finding is that PRISM improves over vanilla SFT across all tested backbones, increasing Factual Avg while also yielding non-negative changes in Common Avg. The effect on Qwen3-4B is comparatively small, which is plausible given the already-strong factual baseline (e.g., HHEM in the mid-90s), suggesting ceiling effects can limit visible headroom. In contrast, gains are clearer for Qwen3-8B and Llama 3.1-8B, where PRISM provides the strongest or near-strongest factual aggregate while remaining competitive on general benchmarks.

The source of Common Avg changes is also informative. For Qwen3-8B, the improvement in the common aggregate is largely driven by higher IFEval scores, while MMLU/GSM8K/HumanEval remain roughly stable across methods. This implies that common-aggregate gains in this setting should be interpreted primarily as better instruction compliance rather than broad improvements in reasoning or coding. Against this backdrop, PRISM’s advantage relative to vanilla SFT is that it improves factuality while maintaining a competitive general profile, rather than trading factual gains for task-specific boosts.

Overall, the baselines and PRISM exhibit complementary behaviors rather than a single uniformly dominant pattern. Compared with vanilla SFT, PRISM is distinguished by its cross-backbone robustness on factual benchmarks without degrading general evaluations, providing context for the ablations in Section~\ref{sec:ablation} on auxiliary strength and update constraints.

\begin{table*}[t]
\centering
\small
\setlength{\tabcolsep}{5pt}
\renewcommand{\arraystretch}{1.12}
\scalebox{0.90}{
\begin{tabular}{l | c c c c | c c c c c}
\toprule
\multirow{2}{*}{\textbf{Variant}} &
\multicolumn{4}{c|}{\textbf{Factual}} &
\multicolumn{5}{c}{\textbf{Common}} \\
\cmidrule(lr){2-5}\cmidrule(lr){6-10}
& \textbf{HHEM} $\uparrow$
& \textbf{PopQA} $\uparrow$
& \textbf{TriviaQA} $\uparrow$
& \textbf{Avg.} $\uparrow$
& \textbf{MMLU} $\uparrow$
& \textbf{GSM8K} $\uparrow$
& \textbf{HumanEval} $\uparrow$
& \textbf{IFEval} $\uparrow$
& \textbf{Avg.} $\uparrow$ \\
\midrule
SFT ($\lambda{=}0$) & 94.53 & 58.54 & 68.95 & 74.01 & 76.19 & 85.14 & 79.27 & 60.81 & 75.35 \\
$\lambda{=}0.01$   & 92.64 & 59.76 & 68.55 & 73.65 & 75.11 & 85.44 & 82.93 & 53.42 & 74.23 \\
$\lambda{=}0.1$    & 95.03 & 59.54 & 68.90 & \textbf{74.49} & 76.93 & 86.20 & 81.71 & 63.40 & \textbf{77.06} \\
$\lambda{=}0.5$    & 94.54 & 58.83 & 68.06 & 73.81 & 74.86 & 86.81 & 79.88 & 51.76 & 73.33 \\
$\lambda{=}1.0$    & 94.54 & 60.33 & 67.44 & 74.10 & 73.12 & 86.05 & 81.71 & 54.53 & 73.85 \\
\bottomrule
\end{tabular}
}
\caption{Effect of the auxiliary weight $\lambda$ on Qwen3-4B. The averaged results suggest a trade-off between factual correction and general capability preservation as the auxiliary signal becomes stronger.}
\label{tab:ablation_coef}
\end{table*}

\begin{table*}[!tb]
\centering
\small
\setlength{\tabcolsep}{5pt}
\renewcommand{\arraystretch}{1.12}
\scalebox{0.90}{
\begin{tabular}{l | c c c c | c c c c c}
\toprule
\multirow{2}{*}{\textbf{Variant}} &
\multicolumn{4}{c|}{\textbf{Factual}} &
\multicolumn{5}{c}{\textbf{Common}} \\
\cmidrule(lr){2-5}\cmidrule(lr){6-10}
& \textbf{HHEM} $\uparrow$
& \textbf{PopQA} $\uparrow$
& \textbf{TriviaQA} $\uparrow$
& \textbf{Avg.} $\uparrow$
& \textbf{MMLU} $\uparrow$
& \textbf{GSM8K} $\uparrow$
& \textbf{HumanEval} $\uparrow$
& \textbf{IFEval} $\uparrow$
& \textbf{Avg.} $\uparrow$ \\
\midrule
Ours & 95.03 & 59.54 & 68.90 & 74.49 & 76.93 & 86.20 & 81.71 & 63.40 & 77.06 \\
w/o model-aware reallocation & 94.83 & 58.54 & 68.14 & 73.84 & 75.54 & 85.22 & 77.44 & 61.74 & 74.99 \\
w/o knowledge mask & 94.04 & 59.97 & 68.97 & 74.33 & 74.68 & 84.38 & 80.49 & 52.31 & 72.97 \\
\bottomrule
\end{tabular}
}
\caption{Component analysis on Qwen3-4B. ``Ours'' uses both model-aware reallocation and knowledge masking. ``w/o model-aware reallocation'' keeps factual masking but removes preference-conditioned reallocation, while ``w/o knowledge mask'' keeps model-aware updates without masking. The combined design yields the best overall common average while remaining competitive on factual benchmarks.}
\label{tab:ablation_modelaware}
\end{table*}

\subsection{Ablation Study}
\label{sec:ablation}

\paragraph{Effect of the compensation weight.}
Table~\ref{tab:ablation_coef} studies the effect of the auxiliary weight $\lambda$ on Qwen3-4B.
Although the behavior is not identical across individual benchmarks, the aggregate pattern is clearer when viewed at the level of factual versus common capability.
As $\lambda$ increases, factual performance generally remains competitive and tends to improve overall, whereas common capability shows a downward trend.
This suggests that stronger probability reallocation shifts the model toward more aggressive factual correction, but at the cost of broader capability preservation.

At the same time, this trend is not strictly monotonic on every benchmark, indicating that the effect of $\lambda$ is better understood as a trade-off than as a uniform gain.
A small-to-moderate auxiliary weight provides the most favorable balance in our setting.
In particular, $\lambda=0.1$ achieves the strongest overall profile, offering clear gains on factuality-related evaluation while avoiding the larger regressions on general benchmarks observed at higher values.

Overall, these results are consistent with the intended role of the complement objective.
When moderately weighted, it acts as a targeted corrective signal.
When assigned too much weight, however, it begins to interfere with the base SFT objective, leading to weaker retention of general reasoning and instruction-following behavior.

\paragraph{Component analysis.}
Table~\ref{tab:ablation_modelaware} compares the full method against two partial variants: one without model-aware reallocation and one without knowledge masking.
The overall pattern suggests that the two components play complementary roles.
Using either component alone can remain competitive on selected benchmarks, but the combined design yields the most balanced profile overall.

Removing model-aware reallocation while retaining knowledge masking leads to a broadly weaker profile, indicating that factual masking alone is not sufficient to fully realize the benefit of the auxiliary signal.
This is consistent with the role of model-aware gating: it makes the update more selective by activating reallocation only when the model is currently inclined toward a risky continuation.
Without this selectivity, the correction becomes less targeted and therefore less effective.

Conversely, removing knowledge masking while keeping model-aware updates preserves competitiveness on some benchmarks, but substantially weakens the overall common-capability profile, with the largest drop appearing on instruction-following.
This suggests that model-aware updates alone are not enough; they still benefit from an explicit factual prior that constrains where correction should be applied.
This comparison suggests a functional division between the two components: knowledge masking provides localization, while model-aware reallocation provides precision.

\section{Conclusion}
\label{sec:conclusion}

We propose PRISM, a training-time method that injects coarse fact structure into supervised fine-tuning via model-aware probability reallocation on fact-critical spans. PRISM augments standard SFT with a lightweight auxiliary objective that selectively penalizes high-confidence learning on weakly supported factual targets, with its scope controlled by span-level risk weights and model-aware gating. Across multiple backbones, PRISM improves factual aggregates while maintaining a competitive profile on general benchmarks covering reasoning, code generation, and instruction following. 

Our ablations further clarify the roles of the two core components. Varying the auxiliary weight reveals a trade-off between stronger factual correction and preservation of broader capabilities, suggesting that the complement objective is most effective when used conservatively. Component analysis further shows that knowledge masking provides \emph{localization}, while model-aware reallocation provides \emph{precision}; their combination yields the most balanced overall profile.

Future work includes improving the reliability of fact extraction and verification, and exploring more adaptive ways to schedule the auxiliary signal across domains and training stages.

\newpage
\bibliographystyle{named}
\bibliography{ijcai26}

@misc{liu-etal-2024-skyworkreward,
      title={Skywork-Reward: Bag of Tricks for Reward Modeling in LLMs}, 
      author={Chris Yuhao Liu and Liang Zeng and Jiacai Liu and Rui Yan and Jujie He and Chaojie Wang and Shuicheng Yan and Yang Liu and Yahui Zhou},
      year={2024},
      eprint={2410.18451},
      archivePrefix={arXiv},
      primaryClass={cs.AI},
      url={https://arxiv.org/abs/2410.18451}, 
}

@inproceedings{mallen-etal-2023-trust,
    title = "When Not to Trust Language Models: Investigating Effectiveness of Parametric and Non-Parametric Memories",
    author = "Mallen, Alex  and
      Asai, Akari  and
      Zhong, Victor  and
      Das, Rajarshi  and
      Khashabi, Daniel  and
      Hajishirzi, Hannaneh",
    editor = "Rogers, Anna  and
      Boyd-Graber, Jordan  and
      Okazaki, Naoaki",
    booktitle = "Proceedings of the 61st Annual Meeting of the Association for Computational Linguistics (Volume 1: Long Papers)",
    month = jul,
    year = "2023",
    address = "Toronto, Canada",
    publisher = "Association for Computational Linguistics",
    url = "https://aclanthology.org/2023.acl-long.546/",
    doi = "10.18653/v1/2023.acl-long.546",
    pages = "9802--9822"
}

@inproceedings{joshi-etal-2017-triviaqa,
    title = "{T}rivia{QA}: A Large Scale Distantly Supervised Challenge Dataset for Reading Comprehension",
    author = "Joshi, Mandar  and
      Choi, Eunsol  and
      Weld, Daniel  and
      Zettlemoyer, Luke",
    editor = "Barzilay, Regina  and
      Kan, Min-Yen",
    booktitle = "Proceedings of the 55th Annual Meeting of the Association for Computational Linguistics (Volume 1: Long Papers)",
    month = jul,
    year = "2017",
    address = "Vancouver, Canada",
    publisher = "Association for Computational Linguistics",
    url = "https://aclanthology.org/P17-1147/",
    doi = "10.18653/v1/P17-1147",
    pages = "1601--1611"
}

@misc{hendrycks2021mmlu,
      title={Measuring Massive Multitask Language Understanding}, 
      author={Dan Hendrycks and Collin Burns and Steven Basart and Andy Zou and Mantas Mazeika and Dawn Song and Jacob Steinhardt},
      year={2021},
      eprint={2009.03300},
      archivePrefix={arXiv},
      primaryClass={cs.CY},
      url={https://arxiv.org/abs/2009.03300}, 
}

@misc{cobbe2021training,
      title={Training Verifiers to Solve Math Word Problems}, 
      author={Karl Cobbe and Vineet Kosaraju and Mohammad Bavarian and Mark Chen and Heewoo Jun and Lukasz Kaiser and Matthias Plappert and Jerry Tworek and Jacob Hilton and Reiichiro Nakano and Christopher Hesse and John Schulman},
      year={2021},
      eprint={2110.14168},
      archivePrefix={arXiv},
      primaryClass={cs.LG},
      url={https://arxiv.org/abs/2110.14168}, 
}

@misc{chen2021evaluating,
      title={Evaluating Large Language Models Trained on Code}, 
      author={Mark Chen and Jerry Tworek and Heewoo Jun and Qiming Yuan and others},
      year={2021},
      eprint={2107.03374},
      archivePrefix={arXiv},
      primaryClass={cs.LG},
      url={https://arxiv.org/abs/2107.03374}, 
}

@misc{zhou2023instructionfollowingevaluationlargelanguage,
      title={Instruction-Following Evaluation for Large Language Models}, 
      author={Jeffrey Zhou and Tianjian Lu and Swaroop Mishra and Siddhartha Brahma and Sujoy Basu and Yi Luan and Denny Zhou and Le Hou},
      year={2023},
      eprint={2311.07911},
      archivePrefix={arXiv},
      primaryClass={cs.CL},
      url={https://arxiv.org/abs/2311.07911}, 
}

@inproceedings{ouyang2022training,
 author = {Ouyang, Long and Wu, Jeffrey and Jiang, Xu and Almeida, Diogo and others},
 booktitle = {Advances in Neural Information Processing Systems},
 editor = {S. Koyejo and S. Mohamed and A. Agarwal and D. Belgrave and K. Cho and A. Oh},
 pages = {27730--27744},
 publisher = {Curran Associates, Inc.},
 title = {Training language models to follow instructions with human feedback},
 url = {https://proceedings.neurips.cc/paper_files/paper/2022/file/b1efde53be364a73914f58805a001731-Paper-Conference.pdf},
 volume = {35},
 year = {2022}
}

@misc{wei2022finetuned,
      title={Finetuned Language Models Are Zero-Shot Learners}, 
      author={Jason Wei and Maarten Bosma and Vincent Y. Zhao and Kelvin Guu and Adams Wei Yu and Brian Lester and Nan Du and Andrew M. Dai and Quoc V. Le},
      year={2022},
      eprint={2109.01652},
      archivePrefix={arXiv},
      primaryClass={cs.CL},
      url={https://arxiv.org/abs/2109.01652}, 
}

@misc{rafailov2023dpo,
      title={Direct Preference Optimization: Your Language Model is Secretly a Reward Model}, 
      author={Rafael Rafailov and Archit Sharma and Eric Mitchell and Stefano Ermon and Christopher D. Manning and Chelsea Finn},
      year={2024},
      eprint={2305.18290},
      archivePrefix={arXiv},
      primaryClass={cs.LG},
      url={https://arxiv.org/abs/2305.18290}, 
}

@inproceedings{maynez-etal-2020-faithfulness,
    title = "On Faithfulness and Factuality in Abstractive Summarization",
    author = "Maynez, Joshua  and
      Narayan, Shashi  and
      Bohnet, Bernd  and
      McDonald, Ryan",
    editor = "Jurafsky, Dan  and
      Chai, Joyce  and
      Schluter, Natalie  and
      Tetreault, Joel",
    booktitle = "Proceedings of the 58th Annual Meeting of the Association for Computational Linguistics",
    month = jul,
    year = "2020",
    address = "Online",
    publisher = "Association for Computational Linguistics",
    url = "https://aclanthology.org/2020.acl-main.173/",
    doi = "10.18653/v1/2020.acl-main.173",
    pages = "1906--1919"
}

@inproceedings{lin-etal-2022-truthfulqa,
    title = "{T}ruthful{QA}: Measuring How Models Mimic Human Falsehoods",
    author = "Lin, Stephanie  and
      Hilton, Jacob  and
      Evans, Owain",
    editor = "Muresan, Smaranda  and
      Nakov, Preslav  and
      Villavicencio, Aline",
    booktitle = "Proceedings of the 60th Annual Meeting of the Association for Computational Linguistics (Volume 1: Long Papers)",
    month = may,
    year = "2022",
    address = "Dublin, Ireland",
    publisher = "Association for Computational Linguistics",
    url = "https://aclanthology.org/2022.acl-long.229/",
    doi = "10.18653/v1/2022.acl-long.229",
    pages = "3214--3252"
}

@inproceedings{min-etal-2023-factscore,
    title = "{FA}ct{S}core: Fine-grained Atomic Evaluation of Factual Precision in Long Form Text Generation",
    author = "Min, Sewon  and
      Krishna, Kalpesh  and
      Lyu, Xinxi  and
      Lewis, Mike  and
      Yih, Wen-tau  and
      Koh, Pang  and
      Iyyer, Mohit  and
      Zettlemoyer, Luke  and
      Hajishirzi, Hannaneh",
    editor = "Bouamor, Houda  and
      Pino, Juan  and
      Bali, Kalika",
    booktitle = "Proceedings of the 2023 Conference on Empirical Methods in Natural Language Processing",
    month = dec,
    year = "2023",
    address = "Singapore",
    publisher = "Association for Computational Linguistics",
    url = "https://aclanthology.org/2023.emnlp-main.741/",
    doi = "10.18653/v1/2023.emnlp-main.741",
    pages = "12076--12100"
}

@misc{wei2024longform,
      title={Long-form factuality in large language models}, 
      author={Jerry Wei and Chengrun Yang and Xinying Song and Yifeng Lu and Nathan Hu and Jie Huang and Dustin Tran and Daiyi Peng and Ruibo Liu and Da Huang and Cosmo Du and Quoc V. Le},
      year={2024},
      eprint={2403.18802},
      archivePrefix={arXiv},
      primaryClass={cs.CL},
      url={https://arxiv.org/abs/2403.18802}, 
}

@article{ji2023survey,
author = {Ji, Ziwei and Lee, Nayeon and Frieske, Rita and Yu, Tiezheng and Su, Dan and Xu, Yan and Ishii, Etsuko and Bang, Ye Jin and Madotto, Andrea and Fung, Pascale},
title = {Survey of Hallucination in Natural Language Generation},
year = {2023},
issue_date = {December 2023},
publisher = {Association for Computing Machinery},
address = {New York, NY, USA},
volume = {55},
number = {12},
issn = {0360-0300},
url = {https://doi.org/10.1145/3571730},
doi = {10.1145/3571730},
journal = {ACM Comput. Surv.},
month = mar,
articleno = {248},
numpages = {38}
}

@article{huang2025survey,
author = {Huang, Lei and Yu, Weijiang and Ma, Weitao and Zhong, Weihong and Feng, Zhangyin and Wang, Haotian and Chen, Qianglong and Peng, Weihua and Feng, Xiaocheng and Qin, Bing and Liu, Ting},
title = {A Survey on Hallucination in Large Language Models: Principles, Taxonomy, Challenges, and Open Questions},
year = {2025},
issue_date = {March 2025},
publisher = {Association for Computing Machinery},
address = {New York, NY, USA},
volume = {43},
number = {2},
issn = {1046-8188},
url = {https://doi.org/10.1145/3703155},
doi = {10.1145/3703155},
journal = {ACM Trans. Inf. Syst.},
month = jan,
articleno = {42},
numpages = {55}
}

@misc{lewis2020rag,
      title={Retrieval-Augmented Generation for Knowledge-Intensive NLP Tasks}, 
      author={Patrick Lewis and Ethan Perez and Aleksandra Piktus and Fabio Petroni and Vladimir Karpukhin and Naman Goyal and Heinrich Küttler and Mike Lewis and Wen-tau Yih and Tim Rocktäschel and Sebastian Riedel and Douwe Kiela},
      year={2021},
      eprint={2005.11401},
      archivePrefix={arXiv},
      primaryClass={cs.CL},
      url={https://arxiv.org/abs/2005.11401}, 
}

@inproceedings{manakul-etal-2023-selfcheckgpt,
    title = "{S}elf{C}heck{GPT}: Zero-Resource Black-Box Hallucination Detection for Generative Large Language Models",
    author = "Manakul, Potsawee  and
      Liusie, Adian  and
      Gales, Mark",
    editor = "Bouamor, Houda  and
      Pino, Juan  and
      Bali, Kalika",
    booktitle = "Proceedings of the 2023 Conference on Empirical Methods in Natural Language Processing",
    month = dec,
    year = "2023",
    address = "Singapore",
    publisher = "Association for Computational Linguistics",
    url = "https://aclanthology.org/2023.emnlp-main.557/",
    doi = "10.18653/v1/2023.emnlp-main.557",
    pages = "9004--9017"
}

@inproceedings{si-etal-2025-aligning,
    title = "Aligning Large Language Models to Follow Instructions and Hallucinate Less via Effective Data Filtering",
    author = "Si, Shuzheng  and
      Zhao, Haozhe  and
      Chen, Gang  and
      Gao, Cheng  and
      Bai, Yuzhuo  and
      Wang, Zhitong  and
      An, Kaikai  and
      Luo, Kangyang  and
      Qian, Chen  and
      Qi, Fanchao  and
      Chang, Baobao  and
      Sun, Maosong",
    editor = "Che, Wanxiang  and
      Nabende, Joyce  and
      Shutova, Ekaterina  and
      Pilehvar, Mohammad Taher",
    booktitle = "Proceedings of the 63rd Annual Meeting of the Association for Computational Linguistics (Volume 1: Long Papers)",
    month = jul,
    year = "2025",
    address = "Vienna, Austria",
    publisher = "Association for Computational Linguistics",
    url = "https://aclanthology.org/2025.acl-long.804/",
    doi = "10.18653/v1/2025.acl-long.804",
    pages = "16469--16488",
    ISBN = "979-8-89176-251-0"
}

@inproceedings{huang-chen-2024-factalign,
    title = "{F}act{A}lign: Long-form Factuality Alignment of Large Language Models",
    author = "Huang, Chao-Wei  and
      Chen, Yun-Nung",
    editor = "Al-Onaizan, Yaser  and
      Bansal, Mohit  and
      Chen, Yun-Nung",
    booktitle = "Findings of the Association for Computational Linguistics: EMNLP 2024",
    month = nov,
    year = "2024",
    address = "Miami, Florida, USA",
    publisher = "Association for Computational Linguistics",
    url = "https://aclanthology.org/2024.findings-emnlp.955/",
    doi = "10.18653/v1/2024.findings-emnlp.955",
    pages = "16363--16375"
}

@inproceedings{gekhman-etal-2024-fine,
    title = "Does Fine-Tuning {LLM}s on New Knowledge Encourage Hallucinations?",
    author = "Gekhman, Zorik  and
      Yona, Gal  and
      Aharoni, Roee  and
      Eyal, Matan  and
      Feder, Amir  and
      Reichart, Roi  and
      Herzig, Jonathan",
    editor = "Al-Onaizan, Yaser  and
      Bansal, Mohit  and
      Chen, Yun-Nung",
    booktitle = "Proceedings of the 2024 Conference on Empirical Methods in Natural Language Processing",
    month = nov,
    year = "2024",
    address = "Miami, Florida, USA",
    publisher = "Association for Computational Linguistics",
    url = "https://aclanthology.org/2024.emnlp-main.444/",
    doi = "10.18653/v1/2024.emnlp-main.444",
    pages = "7765--7784"
}

@inproceedings{huang-etal-2025-alleviating,
    title = "Alleviating Hallucinations from Knowledge Misalignment in Large Language Models via Selective Abstention Learning",
    author = "Huang, Lei  and
      Feng, Xiaocheng  and
      Ma, Weitao  and
      Fan, Yuchun  and
      Feng, Xiachong  and
      Gu, Yuxuan  and
      Ye, Yangfan  and
      Zhao, Liang  and
      Zhong, Weihong  and
      Wang, Baoxin  and
      Wu, Dayong  and
      Hu, Guoping  and
      Kong, Lingpeng  and
      Xiao, Tong  and
      Liu, Ting  and
      Qin, Bing",
    editor = "Che, Wanxiang  and
      Nabende, Joyce  and
      Shutova, Ekaterina  and
      Pilehvar, Mohammad Taher",
    booktitle = "Proceedings of the 63rd Annual Meeting of the Association for Computational Linguistics (Volume 1: Long Papers)",
    month = jul,
    year = "2025",
    address = "Vienna, Austria",
    publisher = "Association for Computational Linguistics",
    url = "https://aclanthology.org/2025.acl-long.1199/",
    doi = "10.18653/v1/2025.acl-long.1199",
    pages = "24564--24579",
    ISBN = "979-8-89176-251-0"
}

@inproceedings{yuan-etal-2025-beyond,
    title = "Beyond Under-Alignment: Atomic Preference Enhanced Factuality Tuning for Large Language Models",
    author = "Yuan, Hongbang  and
      Chen, Yubo  and
      Cao, Pengfei  and
      Jin, Zhuoran  and
      Liu, Kang",
    editor = "Chiruzzo, Luis  and
      Ritter, Alan  and
      Wang, Lu",
    booktitle = "Findings of the Association for Computational Linguistics: NAACL 2025",
    month = apr,
    year = "2025",
    address = "Albuquerque, New Mexico",
    publisher = "Association for Computational Linguistics",
    url = "https://aclanthology.org/2025.findings-naacl.354/",
    doi = "10.18653/v1/2025.findings-naacl.354",
    pages = "6310--6323",
    ISBN = "979-8-89176-195-7"
}

@misc{nguyen2025smoothing,
      title={Smoothing Out Hallucinations: Mitigating LLM Hallucination with Smoothed Knowledge Distillation}, 
      author={Hieu Nguyen and Zihao He and Shoumik Atul Gandre and Ujjwal Pasupulety and Sharanya Kumari Shivakumar and Kristina Lerman},
      year={2025},
      eprint={2502.11306},
      archivePrefix={arXiv},
      primaryClass={cs.CL},
      url={https://arxiv.org/abs/2502.11306}, 
}

@InProceedings{ghosal2024understanding,
  title = 	 {Understanding Finetuning for Factual Knowledge Extraction},
  author =       {Ghosal, Gaurav Rohit and Hashimoto, Tatsunori and Raghunathan, Aditi},
  booktitle = 	 {Proceedings of the 41st International Conference on Machine Learning},
  pages = 	 {15540--15558},
  year = 	 {2024},
  editor = 	 {Salakhutdinov, Ruslan and Kolter, Zico and Heller, Katherine and Weller, Adrian and Oliver, Nuria and Scarlett, Jonathan and Berkenkamp, Felix},
  volume = 	 {235},
  series = 	 {Proceedings of Machine Learning Research},
  month = 	 {21--27 Jul},
  publisher =    {PMLR},
  url = 	 {https://proceedings.mlr.press/v235/ghosal24a.html}
}

@inproceedings{xue-etal-2025-ualign,
    title = "{UA}lign: Leveraging Uncertainty Estimations for Factuality Alignment on Large Language Models",
    author = "Xue, Boyang  and
      Mi, Fei  and
      Zhu, Qi  and
      Wang, Hongru  and
      Wang, Rui  and
      Wang, Sheng  and
      Yu, Erxin  and
      Hu, Xuming  and
      Wong, Kam-Fai",
    editor = "Che, Wanxiang  and
      Nabende, Joyce  and
      Shutova, Ekaterina  and
      Pilehvar, Mohammad Taher",
    booktitle = "Proceedings of the 63rd Annual Meeting of the Association for Computational Linguistics (Volume 1: Long Papers)",
    month = jul,
    year = "2025",
    address = "Vienna, Austria",
    publisher = "Association for Computational Linguistics",
    url = "https://aclanthology.org/2025.acl-long.299/",
    doi = "10.18653/v1/2025.acl-long.299",
    pages = "6002--6024",
    ISBN = "979-8-89176-251-0"
}

@misc{manakul2023selfcheckgpt,
      title={SelfCheckGPT: Zero-Resource Black-Box Hallucination Detection for Generative Large Language Models}, 
      author={Potsawee Manakul and Adian Liusie and Mark J. F. Gales},
      year={2023},
      eprint={2303.08896},
      archivePrefix={arXiv},
      primaryClass={cs.CL},
      url={https://arxiv.org/abs/2303.08896}, 
}

@misc{lin2024flame,
      title={FLAME: Factuality-Aware Alignment for Large Language Models}, 
      author={Sheng-Chieh Lin and Luyu Gao and Barlas Oguz and Wenhan Xiong and Jimmy Lin and Wen-tau Yih and Xilun Chen},
      year={2024},
      eprint={2405.01525},
      archivePrefix={arXiv},
      primaryClass={cs.CL},
      url={https://arxiv.org/abs/2405.01525}, 
}

@misc{zheng2023lmsyschat1m,
      title={LMSYS-Chat-1M: A Large-Scale Real-World LLM Conversation Dataset}, 
      author={Lianmin Zheng and Wei-Lin Chiang and Ying Sheng and Tianle Li and Siyuan Zhuang and Zhanghao Wu and Yonghao Zhuang and Zhuohan Li and Zi Lin and Eric P. Xing and Joseph E. Gonzalez and Ion Stoica and Hao Zhang},
      year={2024},
      eprint={2309.11998},
      archivePrefix={arXiv},
      primaryClass={cs.CL},
      url={https://arxiv.org/abs/2309.11998}, 
}

@misc{openleecher2024lmsyschat1mclean,
  title        = {lmsys\_chat\_1m\_clean},
  author       = {{OpenLeecher}},
  year         = {2024},
  howpublished = {Hugging Face Datasets},
  url          = {https://huggingface.co/datasets/OpenLeecher/lmsys_chat_1m_clean},
  note         = {Initial release July 2024; accessed 2026-01-16}
}

@misc{qwen3technicalreport,
      title={Qwen3 Technical Report}, 
      author={An Yang and Anfeng Li and Baosong Yang and Beichen Zhang and others},
      year={2025},
      eprint={2505.09388},
      archivePrefix={arXiv},
      primaryClass={cs.CL},
      url={https://arxiv.org/abs/2505.09388}, 
}

@misc{llama3-herd,
      title={The Llama 3 Herd of Models}, 
      author={Aaron Grattafiori and Abhimanyu Dubey and Abhinav Jauhri and Abhinav Pandey and others},
      year={2024},
      eprint={2407.21783},
      archivePrefix={arXiv},
      primaryClass={cs.AI},
      url={https://arxiv.org/abs/2407.21783}, 
}

@misc {hhem21open,
    author       = {Miaoran Li and Rogger Luo and Ofer Mendelevitch},
    title        = {{HHEM-2.1-Open}},
    year         = 2024,
    url          = { https://huggingface.co/vectara/hallucination_evaluation_model },
    doi          = { 10.57967/hf/3240 },
    publisher    = { Hugging Face }
}

@inproceedings{
asai2024selfrag,
author={Asai, Akari and Wu, Zeqiu and Wang, Yizhong and Sil, Avirup and Hajishirzi, Hannaneh},
title={Self-{RAG}: Learning to Retrieve, Generate, and Critique through Self-Reflection},
booktitle={The Twelfth International Conference on Learning Representations},
year={2024},
url={https://openreview.net/forum?id=hSyW5go0v8}
}

@misc{izacard2021contriever,
      title={Unsupervised Dense Information Retrieval with Contrastive Learning}, 
      author={Gautier Izacard and Mathilde Caron and Lucas Hosseini and Sebastian Riedel and Piotr Bojanowski and Armand Joulin and Edouard Grave},
      year={2021},
      url = {https://arxiv.org/abs/2112.09118},
      doi = {10.48550/ARXIV.2112.09118},
}

\newpage
\appendix

\section{Implementation Details}

Unless otherwise stated, all methods are trained under a unified SFT setup to enable controlled comparisons. We use one epoch of training with AdamW, learning rate $1\times 10^{-5}$, cosine learning-rate schedule, warmup ratio $0.05$, maximum sequence length $32768$, and global batch size $128$. We employ DeepSpeed ZeRO-2 with CPU offload, BF16 mixed precision, and FlashAttention-2. At evaluation time, we keep the same decoding configuration across methods; for baselines with method-specific inference components, we adapt them to this shared decoding protocol to avoid confounding improvements with stronger search procedures.

\section{More Information}

Table~\ref{tab:data_stats} shows that the preprocessing pipeline produces dense fact-level supervision rather than a small number of coarse instance-level labels. Across 265K verified training instances, we obtain over 3.29M fact items, 867K dependency relations, and 3.14M span annotations, indicating that factual structure is both frequent and non-trivial in the processed data. The 324K user-input-supported facts further show that instruction-grounded support is common, motivating its inclusion in our fact verification and update gating pipeline.

\label{sec::more_info}

\begin{table}[t]
\centering
\small
\setlength{\tabcolsep}{7pt}
\renewcommand{\arraystretch}{1.10}
\begin{tabular}{p{0.72\columnwidth} r}
\toprule
\textbf{Quantity} & \textbf{Count} \\
\midrule
\multicolumn{2}{l}{\textbf{Training instances}} \\
Verified instances retained for training & 265{,}568 \\
\midrule
\multicolumn{2}{l}{\textbf{Fact structure}} \\
Fact items (atomic facts extracted) & 3{,}291{,}756 \\
Fact relations (directed support/dependency edges) & 867{,}202 \\
User-input-supported facts (relations grounded in user input) & 324{,}310 \\
\midrule
\multicolumn{2}{l}{\textbf{Span annotations}} \\
Total span annotations produced & 3{,}136{,}976 \\
Masked chunks (chunks containing $\geq$1 fact item) & 781{,}180 \\
\bottomrule
\end{tabular}
\caption{Summary statistics of the fact-aware preprocessing applied to \textit{lmsys\_chat\_1m\_clean}.}
\label{tab:data_stats}
\end{table}

\end{document}